\documentclass[runningheads]{llncs}
\usepackage{array}
\usepackage{xcolor}
\usepackage{pgfplots}
\usepackage{enumitem}
\usepackage{subfigure} 
\usepackage{tabularx}
\definecolor{myblue}{RGB}{218, 232, 252} 
\definecolor{mygrey}{RGB}{230, 230, 230} 
\usepackage{caption}
\usepackage{subcaption}
\usepackage{textcomp}
\usepackage{amssymb}
\usepackage{balance}
\usepackage{svg}
\usepackage{algorithmic}
\usepackage{amsmath}
\usepackage[linesnumbered, ruled, vlined]{algorithm2e}

\usepackage{amsmath,amsfonts,bm}

\def\eqref#1{equation~\ref{#1}}

\def\1{\bm{1}}

\DeclareMathAlphabet{\mathsfit}{\encodingdefault}{\sfdefault}{m}{sl}
\SetMathAlphabet{\mathsfit}{bold}{\encodingdefault}{\sfdefault}{bx}{n}

\usepackage{tabularray}
\usepackage{makecell}
\usepackage{multirow}
\usepackage[utf8]{inputenc} 
\usepackage[T1]{fontenc}    
\usepackage{threeparttable}
\usepackage{hyperref}       
\usepackage{url}            
\usepackage{booktabs}       
\usepackage{amsfonts}       
\usepackage{nicefrac}       
\usepackage{microtype}      
\usepackage{soul}
\usepackage{colortbl}
\usepackage{booktabs}
\usepackage[misc]{ifsym}
\usepackage{makecell}
\usepackage{wrapfig}
\usepackage{graphicx}
\usepackage{bbm}

\usepackage[T1]{fontenc}
\usepackage{graphicx}
\begin{document}
\title{Knowledge Hierarchy Guided Biological-Medical Dataset Distillation for Domain LLM Training}
\titlerunning{Knowledge Hierarchy Guided Dataset Distillation}
%
\author{Xunxin Cai\inst{1,2}$^\dagger$\and
Chengrui Wang\inst{1}$^\dagger$ \and
Qingqing Long\inst{1} \and\\
Yuanchun Zhou\inst{1,2,3}\and
Meng Xiao\inst{1}$^{(\textrm{\Letter})}$}
\institute{Computer Network Information Center, Chinese Academy of Sciences, Beijing, China
\email{\{xxcai,crwang,qqlong,zyc,shaow\}@cnic.cn}
 \and
University of Chinese Academy of Sciences, Beijing, China
\and
Hangzhou Institute for Advanced Study, University of Chinese Academy of Sciences
}
\def\thefootnote{$\dagger$}\footnotetext{These authors contributed equally to this work.}

\renewcommand\thefootnote{\arabic{footnote}}
%
%
\maketitle              
\begin{abstract}
The rapid advancement of large language models (LLMs) in biological-medical applications has highlighted a gap between their potential and the limited scale and often low quality of available open-source annotated textual datasets. 
In addition, the inherent complexity of the biomedical knowledge hierarchy significantly hampers efforts to bridge this gap.
Can LLMs themselves play a pivotal role in overcoming this limitation?
Motivated by this question, we investigate this challenge in the present study.
We propose a framework that automates the distillation of high-quality textual training data from the extensive scientific literature. 
Our approach self-evaluates and generates questions that are more closely aligned with the biomedical domain, guided by the biomedical knowledge hierarchy through medical subject headings (MeSH). 
This comprehensive framework establishes an automated workflow, thereby eliminating the need for manual intervention. 
Furthermore, we conducted comprehensive experiments to evaluate the impact of our framework-generated data on downstream language models of varying sizes. 
Our approach substantially improves question-answering tasks compared to pre-trained models from the life sciences domain and powerful close-source models represented by GPT-4. 
Notably, the generated AI-Ready dataset enabled the Llama3-70B base model to outperform GPT-4 using MedPrompt with multiple times the number of parameters. 
Detailed case studies and ablation experiments underscore the significance of each component within our framework\footnote{Our code is shared on Github: \href{https://github.com/coco11563/KAILIN-Knowledge-Hierarchy-Guided-Biological-Medical-Dataset-Distillation-for-Domain-LLM-Training}{link}.}.
\end{abstract}

\section{Introduction}

The rise of LLMs has revolutionized bioinformatics, driving the adoption of automated applications across areas~\cite{intro-2-1}, with their effectiveness increasingly validated in real-world Question-Answer (QA)~\cite{intro-2-2}. 
Nevertheless, the inherent complexity of biomedical tasks means that general LLMs often fail to give correct answer unless they are carefully adapted and fine-tuned~\cite{intro-3-3}.
Additionally, the limited availability of substantial biomedical text data impedes the fine-tuning of domain-specific LLMs.
While biomedical research papers indeed serve as rich sources of quality and dependable corpora, they are characterized by complex terminologies and detailed conceptual frameworks that demand considerable human effort for understanding and processing.~\cite{intro-3-2,intro-3-1}. 
Those observations lead to an essential question: \textbf{How to automatically distill high-quality, large-scale datasets from extensive research papers, thus support LLM training?} 

To address the automated corpora distillation challenge, existing frameworks can be categorized primarily into three approaches:
(1) \textit{Predefined rules-based approaches}~\cite{labrak2024biomistral,data_process-1} undertake extensive data cleaning by filtering and standardizing large-scale bioinformatics datasets. 
While those approaches reduce noise and improve data quality, they incur significant operational costs and limit scalability due to the human labor. 
(2) \textit{Knowledge graph-based approaches}~\cite{wu2024pmc,intro-prior-kg} leverage biomedical text data to create comprehensive knowledge structures, but the reliance on curated databases results in inefficiencies and scalability challenges. 
(3) \textit{Synthesis approaches}~\cite{medsyn-intro-prior-related-syn-2,med-text-mining,cai2023resolving} present a promising automated solution to generate question-answer pairs and process large volumes of documents by using LLMs. 
Nonetheless, these studies neglect the integration of cross-disciplinary collaboration~\cite{xiao2023interdisciplinary,intro-isolate-source-bad}, resulting in a lack of diversity and reliability.  

Motivated by those limitations, we propose \textbf{K}nowledge hier\textbf{A}rchy gu\textbf{I}ded biologica\textbf{L}-med\textbf{I}cal dataset distillatio\textbf{N} (\textbf{KAILIN}), an automated framework that integrates knowledge hierarchy and utilizes multiple LLMs as experts for domain QA training corpora extraction. 
The core idea of KAILIN is to introduce a well-established knowledge hierarchy (i.e., Medical Subject Headings (MeSH)~\cite{mesh}) to assess the alignment of the generated `Question-Answer-Context' pair to the domain understanding. 
This framework begins with fine-tuning two LLMs to generate questions from annotated yet scarce open-source datasets. 
After that, the framework retrieves the context from 23 million collected research articles that are most related to the generated questions. 
The better question is determined by evaluating the retrieved contexts and selecting the one with the superior alignment score to the knowledge hierarchy. 
By that, the automated pipeline of generating preference data is present. 
This dataset is designed to train a language model to craft improved questions from unannotated research articles that align more effectively with the existing knowledge structure. 
Using the improved question, we can retrieve the related context, generate answers, and finally form the AI-Ready dataset.

In summary, the key contributions of this work can be summarized as:
\begin{itemize}[label=\textbullet]
    \item \textbf{Biomedical Dataset Distillation Workflow:} We present a comprehensive, highly automated workflow for distilling biomedical corpora from large-scale research articles. This framework enables the creation of expansive, domain-specific training datasets without the need for manual annotation, significantly reducing the cost and time involved in dataset preparation.
    \item \textbf{Framework and Methodology:} We proposed the \textbf{KAILIN} framework which incorporates a MeSH-based knowledge hierarchy similarity evaluation method to integrate and evaluate the quality of the distilled biomedical corpora. 
    KAILIN efficiently constructs high-quality datasets by combining knowledge-based evaluation with context-aware selection, obviating the need for human intervention in dataset curation.
    \item \textbf{Empirical Validation and Insights:} We conducted extensive experiments to validate the effectiveness of our framework and the resulting datasets. 
    Through ablation studies and case analyses, we explored the impact of each technical component. Additionally, we investigated the scaling law of dataset distillation across various settings and model hyperparameter selections. 
\end{itemize}

\section{Methodology}
In this section, we introduce the KAILIN framework, which aims to enhance the open-source dataset through the dataset distillation process, and validate the effectiveness on downstream tasks.

\subsubsection{Fine-Tuning Question Generator:}

To enhance the performance of general-purpose base models in biomedical question generation, we employed biomedical open-source BioASQ dataset\cite{bioasq-1} as the training set $\mathcal{T}$.
Using this training set $\mathcal{T}$, we trained two distinct question generators $\theta^1 $ and $\theta^2$, with LLaMA-2-7B and BioMistral as base models $\theta$ respectively.

\subsubsection{Retrieval Process:}
We collected 23 million abstracts from PubMed\footnote{PubMed:\url{https://pubmed.ncbi.nlm.nih.gov/}}, which serves as the raw dataset $\mathcal{R}$ for dataset distillation and is also constructed as a vector database $\mathcal{V}$ for retrieval purposes. 
Given a document $d_i \in \mathcal{R}$, and two question generation models $\theta^1$ and $\theta^2$ that use $d_i$ as input to infer questions $q^1_i$ and $q^2_i$. We then used these questions $q^1_i$ and $q^2_i$ respectively as input to retrieve the top-$k$ most similar documents from the vector database $\mathcal{V}$. 
These retrieved raw documents, serving as the contexts $c^1_i$ and $c^2_i$ associated with $q^1_i$ and $q^2_i$ respectively, will be used for the subsequent knowledge hierarchy similarity evaluation.


\begin{figure}[h!]
    \centering
    \includegraphics[width=\textwidth]{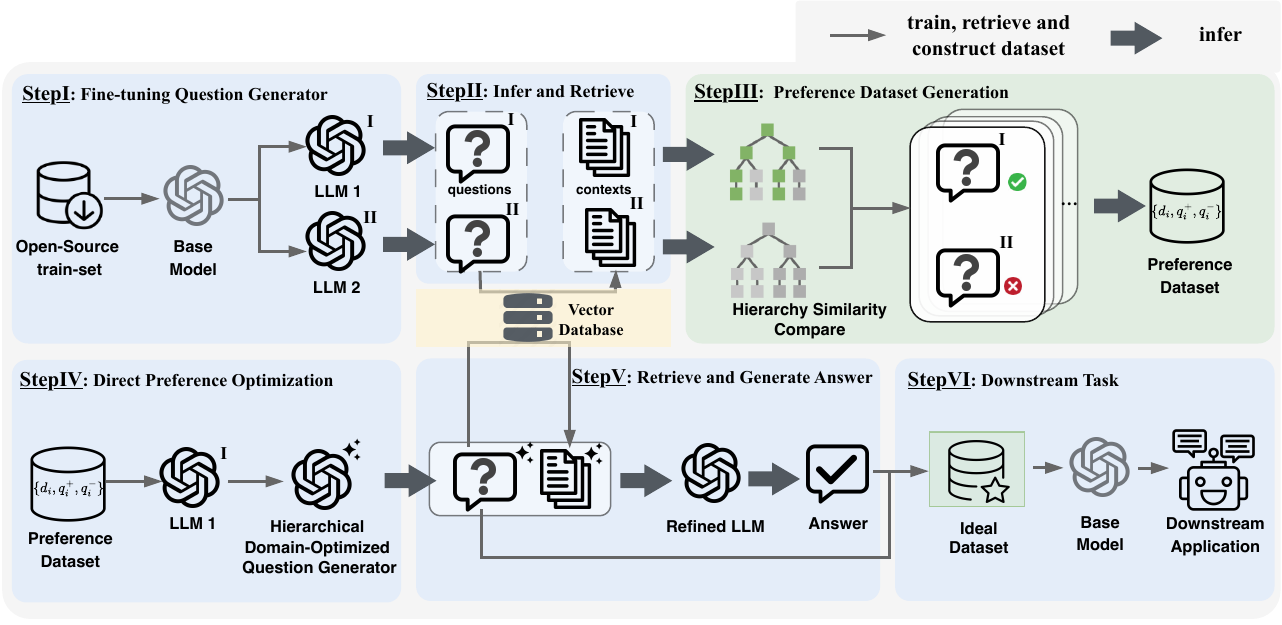} 
    \caption{The overview of KAILIN framework. }
    \label{fig:Fig_overview}
\end{figure}

\subsubsection{Knowledge Hierarchy Similarity Evaluation: } \label{method-Evaluation} 
Using the retrieved contexts $c^1_i$ and $c^2_i$ associated with the generated questions $q^1_i$ and $q^2_i$, we conducted a knowledge hierarchy similarity evaluation between each context and the original document $d_i$ that used for generating the questions. 
This allows us to select the question that better aligns with the knowledge hierarchy of the biomedical field.
For similarity evaluation, we introduced Medical Subject Headings (MeSH), denoted as $\mathcal{M}$. 
MeSH is a hierarchical classification system centered on an well-organized vocabulary that systematically classifies and organizes biomedical knowledge through structured subject terms. 
We present $\prec$, a partial order illustrating the \textit{Belong-to} relationship, to explain how various terms are interrelated. The characteristics of $\prec$ include being asymmetric, anti-reflexive, and transitive\cite{wu2005learning,10052697}:
\vspace{-0.2cm}
\begin{small}
 \begin{align} 
   & \bullet \text{The only one greatest category }\textit{root}\text{ is the root of the } \mathcal{M},  \nonumber\\
   & \bullet \forall m^x_i \in M_i, m^y_j \in M_j, m^x_i \prec m^y_j \to \space m^y_j \not\prec m^x_i,  \nonumber\\
   & \bullet \forall m^x_i \in M_i, m^x_i \not\prec m^x_i, \nonumber\\
   & \bullet \forall m^x_i \in M_i, m^y_j \in M_j, m^z_k \in M_k, m^x_i \prec m^y_j \land m^y_j \prec m^z_k \to m^x_i \prec m^z_k.  \nonumber
   \nonumber
 \end{align}
\end{small} 
Finally, we define the Hierarchical MeSH Structure $\mathcal{M}$ as a partial order set $\mathcal{M}=(\mathbf{M},\prec)$, where $\mathbf{M}=\{M_i\}_{i=1}^{n}$ is a level-organized term set and $n$ denote the total depth. 
We then analyzed the structured subject terms in the contexts $c^1_i, c^2_i$ and the original document $d_i$, incorporating the information content of their hierarchical positions. 
For any structured subject term $m$, we first calculate its information content as:

\begin{equation}
IC(m) = -\log(\frac{freq(\mathcal{M}(m))}{n_{terms}}),
\end{equation}

where $\mathcal{M}(m)$ denotes the set of all descendants of MeSH term $m$, and $n_{terms}$ represents the total number of MeSH terms in the corpus. 
The IC reflects the specificity of a term; rarer terms have higher IC values. 
For any two MeSH terms $m^x$ and $m^y$, we identify their Lowest Common Ancestor (LCA) in the MeSH hierarchy as ${\Lambda}(m^x, m^y)$. 
Referring to Lin's approach\cite{sim_lin}, we calculate the semantic similarity in a taxonomy based on information content as:

\begin{equation}
    S_{x,y} = \frac{2 \times IC({\Lambda}(m^x, m^y))}{IC(m^x) + IC(m^y)},
\end{equation}
where the $S_{x,y}$ represent similarity between $m^x$ and $m^y$. 
We then calculate the final similarity of the knowledge hierarchy between context $c^1_i$ and original document $d_i$ by averaging over all pairwise comparisons between terms as:

\begin{equation}
    \Bar{S}^1_i = \frac{1}{|d_i| |c_i^1|}\sum_{m^x \in d_i}{\sum_{m^y \in c^1_i}}{S_{x, y}},
\end{equation}    
where $\Bar{S}^1_i$ denotes the knowledge hierarchy similarity between original document $d_i$ and associated context $c^1_i$ for question $q^1_i$. 
Similarly, we calculate the knowledge hierarchy similarity $\Bar{S}^2_i$ between the original document $d_i$ and $c^2_i$.

\subsubsection{Preference Dataset Construction.} 
Using the knowledge hierarchy similarity evaluation metrics, we conducted a similarity comparison of the generated questions $q^1_i$ and $q^2_i$ on a large number of original documents $d_i \in \mathcal{R}$. By comparing the corresponding similarity $\Bar{S}^1_i$ and $\Bar{S}^2_i$, we assessed the alignment of $q^1_i$ and $q^2_i$ with the biomedical knowledge hierarchy. Based on such comparison, we consider $q^1_i$ or $q^2_i$ with the higher similarity score to be better aligned and designate it as $q^+_i$, while the other is designated as $q^-_i$ and constructed a preference dataset $\mathcal{P} = \{d_i, q_i^+, q_i^-\}_{i=1}^N$.

\subsubsection{Direct Preference Optimization.} 
With the preference data pairs prepared offline as $\mathcal{P}$, we employed direct preference optimization (DPO)\cite{DPO} for model alignment. 
We performed DPO to get optimized question generator $\theta_3$, $\theta_1 \xrightarrow{\mathcal{P}} \theta^3$. 

\subsubsection{Ideal Dataset Construction}
We employed the optimized question generation model $\theta_3$ to utilize a original document $d_j \in \mathcal{R}$ as input and generate optimized question $q_j$. Documents related to $q_j$ were then retrieved as context $c_j$, and we further employed the LLaMA-3-70B and GPT-4o to generate ideal answers $a_j$. 
We constructed two different ideal datasets for distinct purposes. 
To enhance the model's foundational understanding in the biomedical field, we combined questions $q_j$ and relevant contexts $c_j$ to form an ideal dataset $ \mathcal{I}_1 = \{q_j, c_j\}_{j=1}^N$ for continued pre-training. 
For improving question-answering performance, we combined questions $q_j$, relevant contexts $c_j$, and corresponding answers $a_j$ to form an ideal supervised fine-tuning dataset $\mathcal{I}_2 = \{q_j, c_j, a_j\}_{j=1}^N$.

\subsubsection{Training for Downstream Task}
When we proceeded with further training for biomedical question-answering applications, we utilized the $\mathcal{I}_1$ dataset for continued pretraining to achieve better performance improvements along with the subsequent supervised fine-tuning. 
We describe the two training stages in detail as:
\begin{itemize} [label=\textbullet]

    \item \textbf{Continuous Pre-training.} We utilize heuristic questions $q_j$ generated by the KAILIN framework, along with the retrieved Top-$k$ documents $c_j$ associated with them, as the corpus for continuous pre-training, using the prompt.  
    With the pre-training corpus constructed in this manner, we continued to pre-train our base models with varying parameter sizes. 

    \item \textbf{Supervised Fine-tuning.} To optimize the model's question-answering performance following continuous pre-training, we further performed full-parameter fine-tuning on the models with the PQA-A training set from PubMedQA, with the prompt. 

\end{itemize}



\section{Experiment}
\subsection{Experimental Setups} 

\textbf{Base Models and Baselines.} We utilized Llama-2-7B, Llama-2-13B~\cite{touvron2023llama}, Llama-3-8B, and Llama-3-70B~\cite{llama3_1} as the base models in our primary experiment, while also incorporating BioMistral~\cite{labrak2024biomistral} for building the preference dataset in training the question generator. 
We conducted a comprehensive evaluation of various open-source models, including LLaMA-2~\cite{touvron2023llama}, LLaMA-3~\cite{llama3_1}, Mistral~\cite{jiang2023mistral}, and Gemma\cite{team2024gemma}, as well as proprietary models like GPT-4\cite{achiam2023gpt}, and PaLM~\cite{Med-Palm-2}. In particular, we focused on models specifically trained for the biomedical domain, such as BioMistral~\cite{labrak2024biomistral}, PMC-LLaMA~\cite{wu2024pmc}, HEAL~\cite{HEAL}, and MMedLM~\cite{MMed_LM}, to demonstrate the effectiveness of our approach. 

\subsubsection{Evaluation Datasets.} We validated the results of our main experiment using the PubMedQA benchmark~\cite{jin2019pubmedqa}, a dataset specifically designed to assess the performance of question-answering systems in the biomedical domain. PubMedQA is tailored to address questions relevant to biomedical literature, making it highly suitable for assessing our framework's adaption in this field. Additionally, we categorized the benchmark based on Medical Subject Headings (MeSH)~\cite{mesh} and publication dates, enabling us to evaluate our system's improved understanding and robustness across diverse biomedical topics and varying time spans.

\begin{figure}[t]
    \centering
    \subfigure[LLMs of fewer than 13B parameters]{
        \includegraphics[width=0.45\textwidth]{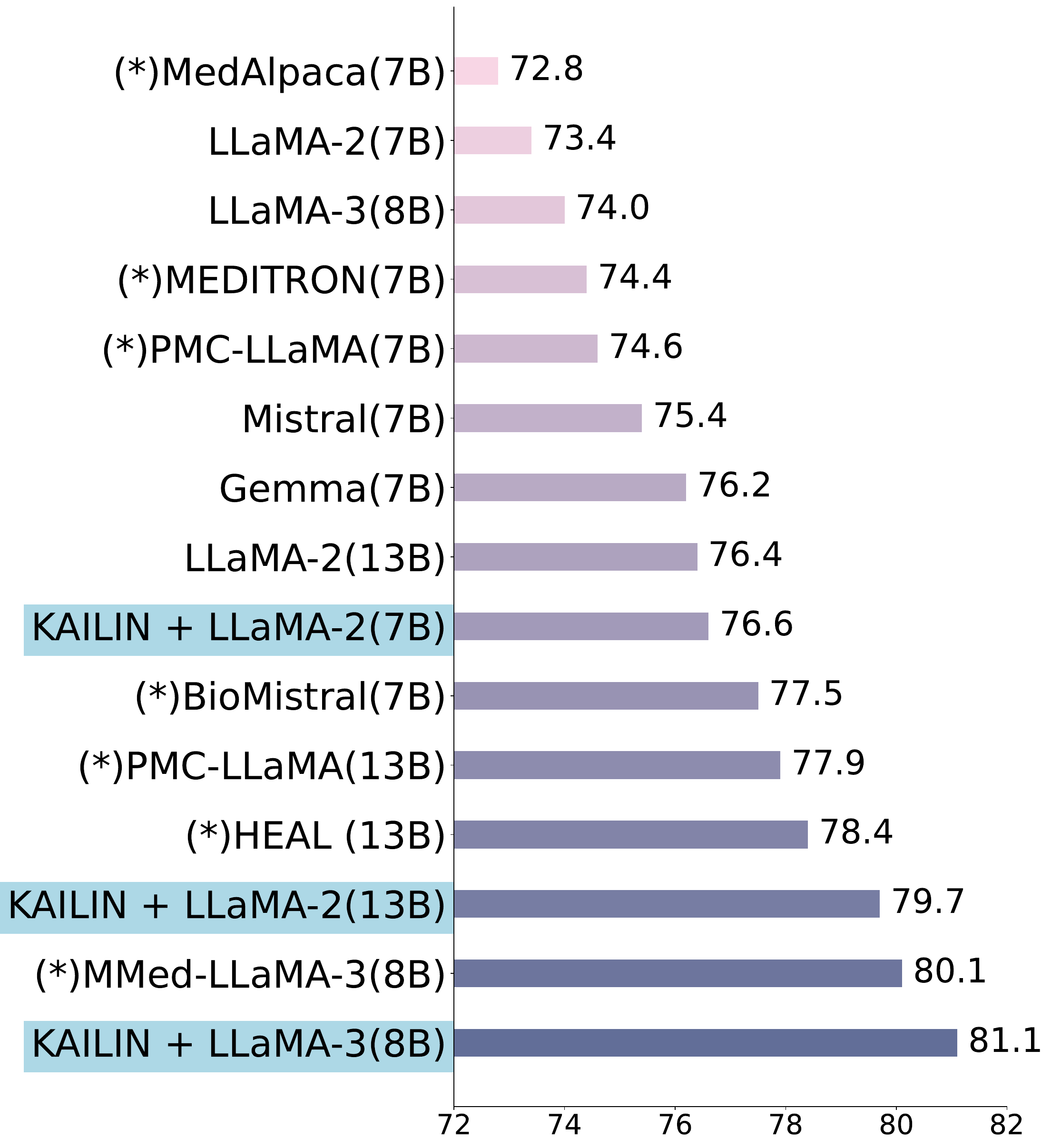} 
        \label{fig:sub1}
    }
    \hspace{0.0\textwidth} 
    \subfigure[LLMs of more than 70B parameters]{
        \includegraphics[width=0.45\textwidth]{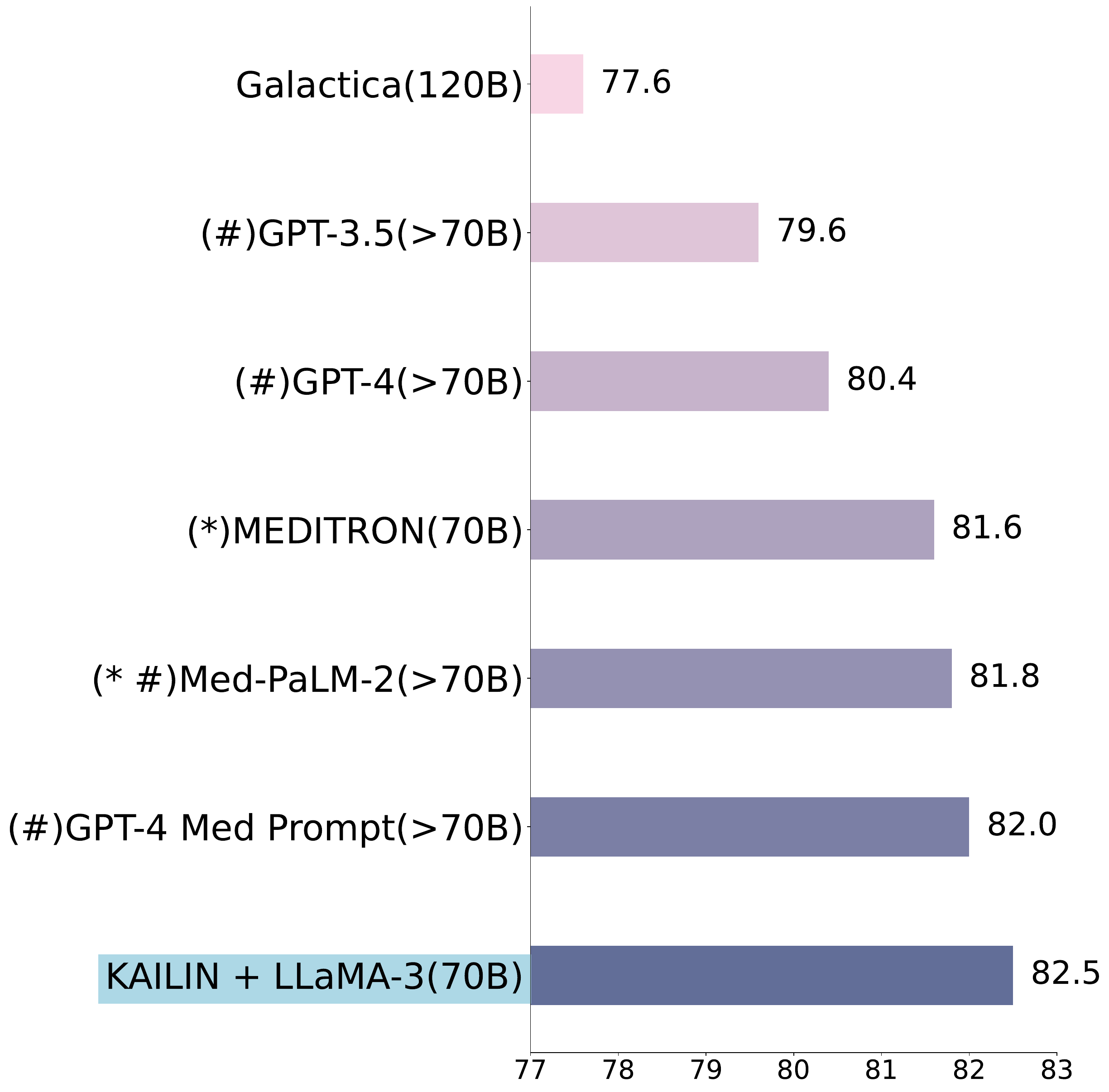}
        \label{fig:sub2}
    }
    \caption{Evaluations (accuracy (\%)) for PubMedQA~\cite{jin2019pubmedqa} problems on our models compared to other open-source models, closed-source models, and domain-specific models. Models marked with * indicate that they are domain-specific large models focused on the biomedical field rather than general-purpose models. The \# symbol denotes closed-source models as opposed to open-source ones.}
    \label{main_exp}
\end{figure}

\subsection{Main Results and Analysis} \label{exp-main}

We compared KAILIN against various models divided into two groups based on parameter size: those with fewer than 13B parameters and those with 70B parameters or more. 
When comparing domain-specific models with general-purpose models within LLMs with fewer than 13B parameters, as shown in Figure \ref{main_exp}, we observed that general-purpose models tend to struggle more to excel in domain-specific tasks. 
However, the KAILIN framework enables general-purpose models to outperform domain-specific models with higher training costs, even with minimal additional training. 
The underlying driver stems from the KAILIN framework's use of MeSH-based knowledge hierarchy similarity evaluation, which effectively addresses the comprehension challenges posed by the rich terminologies and complex conceptual structures inherent in biomedical texts. 
This phenomenon indicates that the KAILIN framework excels in training small general-purpose models to adapt more effectively to specific domains, which is particularly advantageous in the fast-paced evolution of large model iterations.

When comparing LLMs with more than 70B parameters, we observed that the KAILIN framework enables LLaMA-3-70B to outperform closed-source models with significantly larger parameter counts, such as GPT-4 with MedPrompt\cite{nori2023can,nori2023capabilities} and Med-PaLM-2\cite{Med-Palm-2}. 
While larger models typically demonstrate superior performance, the KAILIN framework leverages MeSH-based knowledge hierarchy similarity evaluation for preference alignment. 
This approach acts as a pivotal underlying driver, enabling the model to excel in specific tasks and surpass significantly larger counterparts. 
This phenomenon also highlights a potential future direction for training domain-specific models: KAILIN demonstrates the use of smaller datasets that retain a comprehensive understanding of the knowledge hierarchy to optimize performance on domain-specific tasks. 

\subsection{Ablation Study} \label{exp-ablation}

\begin{table}
\centering
\caption{Evaluations (accuracy (\%)) of the overall experimental results of the ablation study. The reasoning-required and question-only are both inference settings in PubMedQA study. }
\resizebox{\textwidth}{!}{

\begin{tabularx}{\textwidth}{l *{3}{>{\centering\arraybackslash}X}}

\toprule
\textbf{} & \textbf{Reasoning-required} & \textbf{Question-only} \\ 

\midrule

w/o \textbf{MeSH} & 69 & 56.4 \\ 

w/o \textbf{Embedding} & 71.8 & 55.6 \\ 

w/o \textbf{Both} & 64.8 & 43 \\ 

\textbf{Full} & \textbf{72.4} & \textbf{57.8} \\ 

\bottomrule

\end{tabularx}
\vspace{-0.6cm}

}

\label{table_exp_ablation}

\end{table}

We investigated the impact of our MeSH-based knowledge hierarchy similarity evaluation by instead utilizing a Term Frequency-Inverse Document Frequency (TF-IDF) approach for evaluating the document collection.
We also investigated the influence of the embedding model during retrieval process by randomly sampling top-$k$ documents and analyzing their replacement in the MeSH-based preference selection process.

As shown in Table \ref{table_exp_ablation}, we found that the overall performance was worse under the ablation setting without MeSH than without Embedding model, highlighting the critical role of MeSH as part of the framework. 
Without MeSH serving as a structured marker for each document within the knowledge hierarchy, the model struggled to align with the overall knowledge hierarchy of the biomedical field through isolated documents. 

\section{Conclusion}
In this paper, we proposed a novel automated dataset distillation framework, namely KAILIN, which integrates domain-specific knowledge hierarchy. 
We aim for KAILIN to serve as an automated approach that distlls datasets of high-quality from existing unlabeled datasets at lower costs while preserving the integrity of domain knowledge hierarchy. 
Our method was evaluated on the PubMedQA leaderboard, and we further assessed performance  robustness across subsets divided by time span and disciplines. 
These evaluations revealed how LLMs perform differently across various time periods and subfields, demonstrating the superiority of our approach. 


\section{Ackownledgement}
This work is partially supported by the  Beijing Natural Science Foundation (No.4254089), the Postdoctoral Fellowship Program of CPSF (No.GZC20232736), the China Postdoctoral Science Foundation Funded Project (No.2023M743565), and National Natural Science Foundation of China (No.92470204).

%

\bibliographystyle{splncs04}
\bibliography{ref}

\end{document}